\documentclass[conference]{IEEEtran}
\IEEEoverridecommandlockouts
\usepackage{cite}
\usepackage{amsmath,amssymb,amsfonts}
\usepackage{algorithmic}
\usepackage{graphicx}
\usepackage{textcomp}
\usepackage{xcolor}
\usepackage[colorlinks=true,linkcolor=blue,urlcolor=blue,citecolor=blue,anchorcolor=blue]{hyperref} 
\usepackage{multirow} 
\usepackage[caption=false]{subfig} 
\usepackage{nicematrix}
\usepackage{booktabs}
\usepackage{pgfplots}
\usepackage{pgfplotstable}
\usepackage{dsfont}
\usepackage{siunitx} 
\usepackage{balance} 

\usetikzlibrary{matrix}

\def\BibTeX{{\rm B\kern-.05em{\sc i\kern-.025em b}\kern-.08em
    T\kern-.1667em\lower.7ex\hbox{E}\kern-.125emX}}

\begin{document}

\title{Continual Learning for Robust Gate Detection under Dynamic Lighting in Autonomous Drone Racing
\thanks{This research was partially supported by the Aarhus University Research Foundation and Independent Research Fund Denmark, DFF-Research Project~1, with case number: 2035-00052B. This research is part of the programme DesCartes and is supported by the National Research Foundation, Prime Minister’s Office, Singapore under its Campus for Research Excellence and Technological Enterprise~(CREATE) programme.}
}

\author{%
\IEEEauthorblockN{Zhongzheng Qiao}
\IEEEauthorblockA{\textit{ERI@N, Interdisciplinary Graduate Programme} \\
\textit{Nanyang Technological University} \\
Singapore \\
qiao0020@e.ntu.edu.sg}
\and
\IEEEauthorblockN{Xuan Huy Pham}
\IEEEauthorblockA{\textit{Department of Electrical and Computer Engineering} \\
\textit{Aarhus University}\\
Aarhus, Denmark \\
huy.pham@ece.au.dk}
\and
\IEEEauthorblockN{Savitha Ramasamy}
\IEEEauthorblockA{\textit{Institute of Infocomm Research~(I2R)} \\
\textit{~~~~~~~~~~Agency for Science, Technology and Research~(A*STAR)}\\
Singapore \\
ramasamysa@i2r.a-star.edu.sg}
\and
\IEEEauthorblockN{Xudong Jiang}
\IEEEauthorblockA{\textit{School of Electrical and Electronic Engineering} \\
\textit{Nanyang Technological University}\\
Singapore \\
exdjiang@ntu.edu.sg}
\and
\IEEEauthorblockN{Erdal Kayacan}
\IEEEauthorblockA{\textit{~~~~~~~~~~~~~~~~~~~~Automatic Control Group (RAT)~~~~~~~~~~~~~~~~~~~~} \\
\textit{Paderborn University}\\
Paderborn, Germany \\
erdal.kayacan@uni-paderborn.de}
\and
\IEEEauthorblockN{Andriy Sarabakha}
\IEEEauthorblockA{\textit{Department of Electrical and Computer Engineering} \\
\textit{Aarhus University}\\
Aarhus, Denmark \\
andriy@ece.au.dk}
}



\maketitle

\begin{abstract}
In autonomous and mobile robotics, a principal challenge is resilient real-time environmental perception, particularly in situations characterized by unknown and dynamic elements, as exemplified in the context of autonomous drone racing. This study introduces a perception technique for detecting drone racing gates under illumination variations, which is common during high-speed drone flights. The proposed technique relies upon a lightweight neural network backbone augmented with capabilities for continual learning. The envisaged approach amalgamates predictions of the gates' positional coordinates, distance, and orientation, encapsulating them into a cohesive pose tuple. A comprehensive number of tests serve to underscore the efficacy of this approach in confronting diverse and challenging scenarios, specifically those involving variable lighting conditions. The proposed methodology exhibits notable robustness in the face of illumination variations, thereby substantiating its effectiveness.
\end{abstract}

\begin{IEEEkeywords}
continual learning, aerial robotics, machine perception
\end{IEEEkeywords}


\section{Introduction}
\label{sec:introduction}

Autonomous drone racing stands as a catalyst for innovation at the confluence of aerial robotics and artificial intelligence~\cite{Kaufmann2023Nature}. This competition-like showcase serves as a real-world test ground where the drones must navigate intricate tracks with high precision and agility. 
Autonomous drone racing presents a myriad of challenges that demand innovative solutions in the domain of robotics and machine learning. Navigating racing tracks at high speeds requires advanced computer vision algorithms to detect the racing gates under rapidly changing illumination along the racing track. This distinctive characteristic makes gate detection one of the most challenging tasks within autonomous drone racing.

\begin{figure}[!t]
    \centering
    \includegraphics[width=\columnwidth]{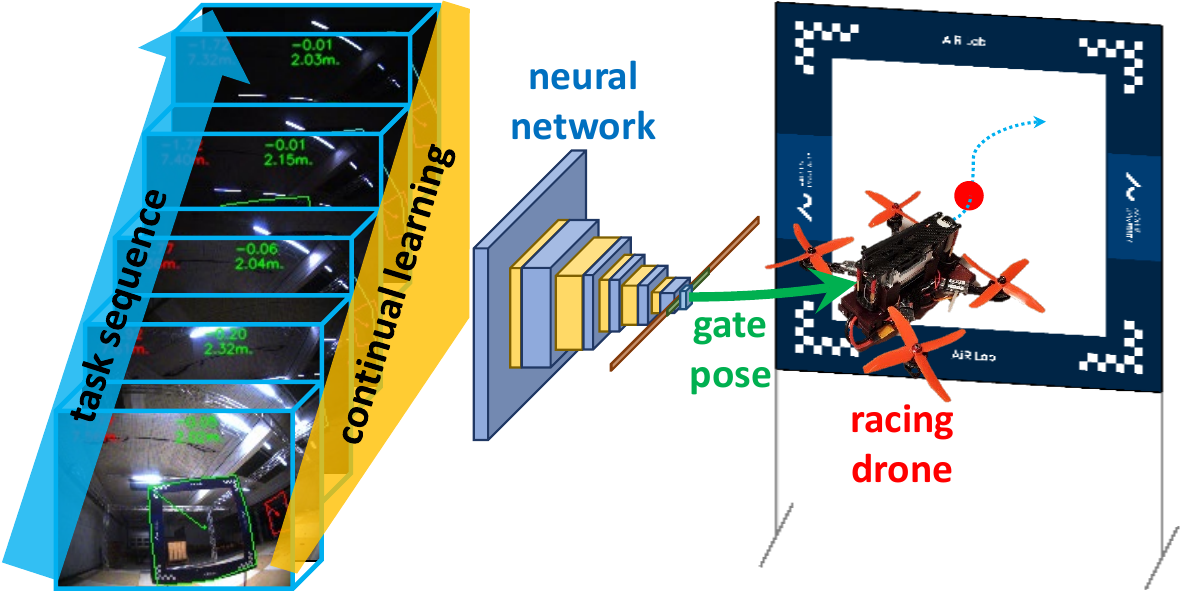}
    \caption{Illustration of the operational principles of the gate perception framework. Images of the racing track are captured using a single fish-eye camera under various illumination conditions. The proposed framework incorporates continual learning methodologies, whereby the neural network undergoes incremental training to detect the gates in various conditions without forgetting any previous conditions.} 
    \label{fig:abstract}
\end{figure}

Early works~\cite{jung2018direct, li2020autonomous, rojas2017metric} commonly employed hand-crafted gate perception layered upon conventional control and planning frameworks to facilitate secure gate passage. However, these approaches exhibit diminished resilience in real-world conditions~\cite{de2021artificial}. Subsequent research has harnessed the power of deep neural networks~(DNNs), proving more robust in dealing with uncertainties inherent in perception tasks~\cite{Morales2020IJCNN}. In~\cite{kaufmann2019beauty}, a ResNet-8 DNN variant with three residual blocks is introduced to estimate gate distributions explicitly. In~\cite{foehn2020alphapilot}, a variation of \mbox{U-Net} is employed for gate segmentation, demonstrating robust performance within a well-engineered framework. Similarly, the approach outlined in~\cite{de2021artificial} also utilizes U-Net for gate segmentation but adopts a unique strategy for associating gate corners, combining corner pixel searching with re-projection error rejection. The study outlined in~\cite{Huy2021IROS} introduced a DNN designed to facilitate resilient gate pose estimation by using raw fisheye RGB images. Subsequently, the research in~\cite{Huy2022RAL} proposed a DNN that leverages morphological operations to enhance the robustness of the gate perception system, particularly in environments characterized by extreme illumination conditions. Recent advancements in this domain~\mbox{\cite{loquercio2018dronet, song2021autonomous, pfeiffer2022visual}}, have explored end-to-end methodologies, employing a singular DNN to generate tracking targets for controllers. The efficacy of gate perception, whether achieved through explicit gate mapping or implicit feature-based comprehension, significantly influences the overall system performance.

On the other hand, continual learning (CL) proves to be a powerful tool for enhancing object detection under incrementally variable conditions~\cite{Menezes2023NN}. Traditional object detection models may encounter difficulties in generalizing across a spectrum of diverse lighting scenarios. In contrast, CL empowers models to dynamically adapt to evolving environments without succumbing to catastrophic forgetting. Through incrementally updating their knowledge base, the models trained with CL can refine the detection of the objects under different lighting conditions, gradually improving accuracy and robustness. Continual learning facilitates the adaptation to new illuminations, preventing the need for constant retraining from scratch. 
The ability to accumulate knowledge over time allows object detection systems to maintain high performance, making them well-suited for autonomous drone racing.

This study introduces a CL approach showcasing an effective learning capability for gate perception in the context of autonomous drone racing navigation, as illustrated in Fig.~\ref{fig:abstract}. Our methodology builds upon the architectural foundation outlined in~\cite{Huy2022RAL} but distinguishes itself by adopting CL instead of conventional sequential learning to train the network across diverse lighting conditions. Notably, our proposed framework eliminates the need for the storage of and access to all training samples, relying solely on samples belonging to a specific task. In this case, the tasks are categorised based on the illumination on the racing track. Experimental results demonstrate robust detection of the gates across dynamic lighting conditions.

This manuscript is organised as follows. Section~\ref{sec:problem} formulates the problem of gate detection within the context of autonomous drone racing. Section~\ref{sec:method} elucidates the proposed methods, which hinge on CL. Experimental results validating the efficacy of the proposed methods are presented in Section~\ref{sec:results}. The concluding remarks and avenues for potential future work are presented in Section~\ref{sec:conclusions}.


\section{Problem Statement}
\label{sec:problem}

The challenge of object detection entails predicting the 3D pose of specific objects in unknown space, subject to high speed under visual degrading conditions. In the context of autonomous drone racing, these objects are referred to as racing gates, in which a drone is required to traverse swiftly without collision to maintain its position in the race. While racing regulations typically specify the predetermined sequence of gates to be traversed, the conventional practice prioritizes the nearest gate for passage.


\begin{figure}[!b]
    \centering
    \includegraphics[width=\columnwidth]{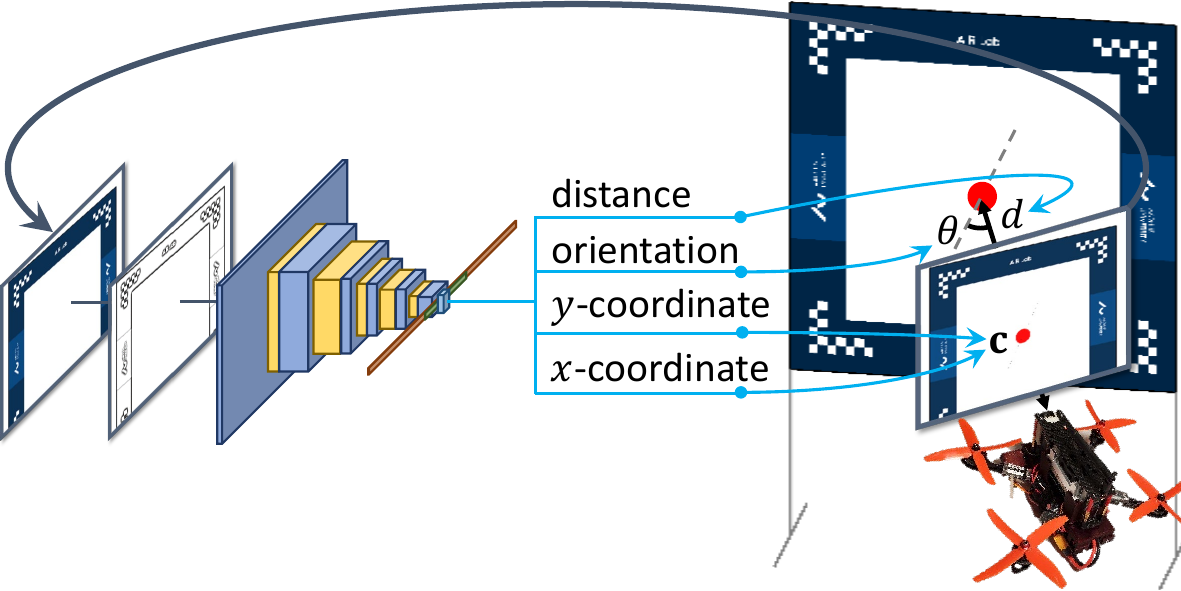}
    \caption{Illustration of the gate perception pipeline. A raw RGB image is converted by the pencil filter before inputting into the neural network, which predicts the center of the gate in the image frame, along with the corresponding distance and orientation of the gate relative to the drone's body frame. Using this information, a 3D pose of the gate can be reconstructed via backprojection.} 
    \label{fig:perception}
\end{figure}

Assuming the gates can be rotated only around their vertical axis, the pose of a gate can be completely described by four parameters: $(x, y, z)$ for 3D-position and $\theta$ for orientation.
Given a raw image frame obtained by the monocular RGB camera, the perception model of the autonomous racing drone has to identify the collection of all of the visible gates on the image. For each gate $i$, visible in the image, the position of the gate's center $\mathbf{c}_i = (x_i, y_i)$, the distance $d_i$ from the drone toward each of the gates' center, and their relative orientation $\theta_i$. From these predicted terms, the poses of the gates $(p_1^W, p_2^W, ..., p_n^W)$ in a global world frame $W$ can be retrieved using back-projection, as illustrated in Fig.~\ref{fig:perception}. A global map can be constructed and utilized by a path planner to fly the robot through the sequence of the gates.

\begin{figure*}[!t]
    \centering
    \includegraphics[width=\textwidth]{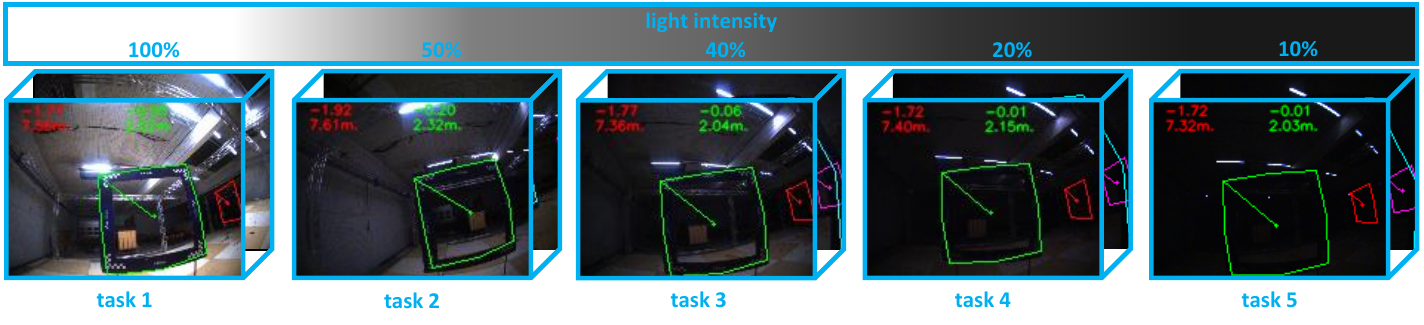}
    \caption{Samples from datasets with environmental illumination varying from $100\%$ to $10\%$ of light intensity.} 
    \label{fig:light_intensity}
\end{figure*}

\begin{figure*}[!t]
    \centering
    \includegraphics[width=\textwidth]{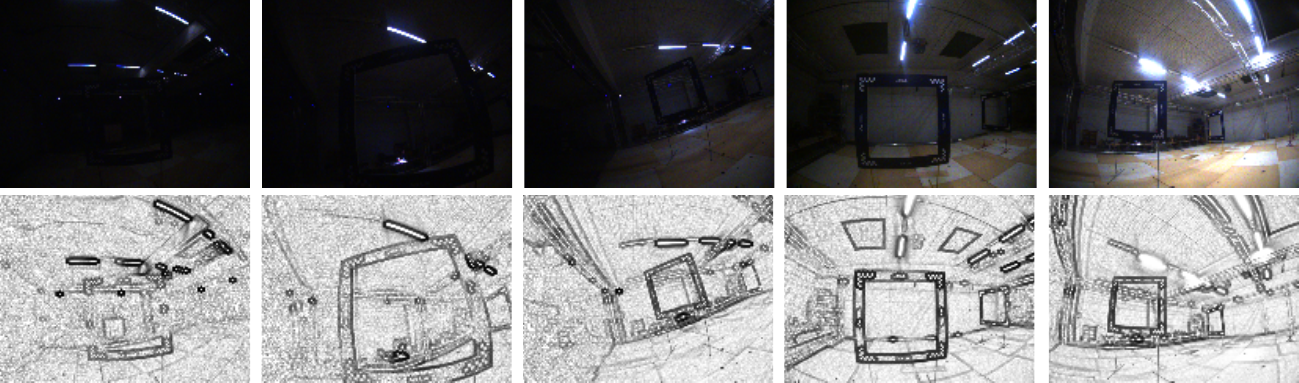}
    \caption{Samples from datasets. Top row: RGB images with illumination varying from $10\%$ to $100\%$.  Bottom row: images after applying pencil filter.} 
    \label{fig:data_samples}
\end{figure*}

A primary challenge is to ensure the robustness of the perception model during the racing scenario, as drone racing is subjected to visual degradation conditions, such as motion blur at high speed, and the changing illumination deliberately introduced during the racing contest. While addressing motion blur often entails reconsidering the entire paradigm of visual image capture and processing for effective robot navigation, our current emphasis is on addressing the impact of changing illumination. To effectively cope with lighting changes, the perception model must learn the core features of the gates under various illumination levels.


\section{Proposed Method}
\label{sec:method}

\subsection{Data Collection}

For this study, real-world data of an autonomous drone completing various racing tracks have been collected. Each data point consists of a raw RGB image associated with the ground-truth pose of the drones and all four gates' pose. From this information, the ground-truth label of each visible gate center, relative distance and relative orientation toward the drone can be computed for each image, as depicted in Fig.~\ref{fig:light_intensity}. To resemble the light intensity challenge in autonomous drone racing, the data are collected in different illumination levels by carefully controlling the lab lighting at night. Thus, $6\,760$ data samples are obtained in five illumination levels: $100 \%$ (full), $50\%$, $40\%$, $20\%$ and $10\%$. Samples of the data and label are visualized in Fig.~\ref{fig:light_intensity}. 

During training, the ground truth is prepared as a 3D tensor with a shape of $R \times C \times 5$, where $R = 4$ and $C = 3$ are the numbers of rows and columns that divide the image into a number of smaller patches. Each patch contains a tuple $\{g, x, y, d, \theta\}$, where $x$ and $y$ are the pixel location of the gate center in the image frame, $d$ is the relative distance from the gate center to the drone, and $\theta$ is the relative orientation of the gate compared to the drone. If the gate center resides in the patch, the confidence value $g$ is set to $1$, otherwise it is $0$. Similarly to~\cite{Huy2022RAL}, the raw images are applied to a morphological filter called "pencil filter", which provides an input abstraction proven to improve the robustness of perception models in poor lighting conditions~\cite{Huy2022RAL}. As can be seen from Fig. \ref{fig:data_samples}, this filter emphasizes the geometrical features of the gates, even in extremely poor light conditions.


\subsection{Gate Detection}


Similarly to~\cite{Huy2021IROS}, PencilNet uses a combination of center coordinate, distance, orientation and confidence losses $\mathcal{L}_{xy}$, $\mathcal{L}_{d}$, $\mathcal{L}_{\theta}$ and $\mathcal{L}_{c}$ respectively compared to their ground-truth labels, that can be described as follows:
\begin{equation}
    \centering
    \begin{cases}
        \mathcal{L}_{xy} &= { {
            \sum_{i = 1}^{R}
                \sum_{j = 1}^{C}      
                  {\mathds{1}}_{ij}^{\text{obj}}
            \left[
            \left(
                x_{ij} - \hat{x}_{ij}
            \right)^2 +
            \left(
                y_{ij} - \hat{y}_{ij}
            \right)^2
            \right]
        }}\\
        \mathcal{L}_{d} &= { {
            \sum_{i = 1}^{R}
                \sum_{j = 1}^{C}      
            {\mathds{1}}_{ij}^{\text{obj}} 
            \left(
                d_{ij} - \hat{d}_{ij}
            \right)^2
        }}\\
        \mathcal{L}_{\theta} &= { {
            \sum_{i = 1}^{R}
                \sum_{j = 1}^{C}      
            {\mathds{1}}_{ij}^{\text{obj}}
            \left(
                \theta_{ij} - \hat{\theta}_{ij}
            \right)^2
        }}\\
        \mathcal{L}_{c} &= { {
            \sum_{i = 1}^{R}
                \sum_{j = 1}^{C}      
                  \left[
                  {\mathds{1}}_{ij}^{\text{obj}}+\alpha (1-{\mathds{1}}_{ij}^{\text{obj}})\right]
            \left(
                c_{ij} - \hat{c}_{ij}
            \right)^2
        }}\\
    \end{cases},
    \label{eq:loss_function}
\end{equation}
where $i\in[1,R]$ and $j\in[1,C]$. For each grid cell of the output layer, the losses are calculated with row $i$ and column $j$. Terms with the hat operator $(\hat{\star})$ denote predicted values. 
The term ${\mathds{1}}_{ij}^{\text{obj}}$ is a binary variable to penalize the network only when a gate center locates in a particular grid. The confidence values for grid cells, where a gate center is not present, are minimized with a weight $\alpha$ so that we can threshold the low confidence predictions in run time, thus reducing false positives. Finally, the loss function for gate detection is a weighted term:
 \begin{equation}
\centering
 \mathcal{L}_{gate} = \lambda_{xy} \mathcal{L}_{xy} + \lambda_{d} \mathcal{L}_{d} + \lambda_{\theta} \mathcal{L}_{\theta} + \lambda_{c}\mathcal{L}_{c},
 \end{equation}
 where $\lambda_{xy}$, $\lambda_{d}$, $\lambda_{\theta}$, $\lambda_{c}$ are weights reflecting importance for each of the losses.

\subsection{Continual Learning}


\subsubsection{Concepts and motivation}
Continual learning has attracted growing attention due to its vital role in adapting neural networks to dynamic environments \cite{de2021continual, wang2023comprehensive}. Different from conventional offline learning, which assumes the availability of a static dataset before training, CL encompasses a paradigm wherein data continuously emerges over time, often exhibiting varying distributions as a consequence of environmental changes. A salient challenge in this incremental learning process is catastrophic forgetting \cite{mccloskey1989catastrophic}, i.e., the model forgets the previously acquired knowledge after learning a new task. CL aims to overcome catastrophic forgetting and facilitate the adaption of new data, leading to a balanced overall performance.
 
Continual learning of drone racing is a practical and crucial issue. In general, the model is initially trained in a controlled base environment, such as an indoor laboratory. However, practical applications often demand the operation of drones in novel and challenging environments, such as outdoor terrains under poor weather conditions. The model needs to be adapted to the new environment efficiently without severe degradation in performance in previous environments. While CL has been explored within the domain of robotics \cite{lesort2020continual}, its application to gate detection in the context of drone racing remains an area that has not been extensively studied. To the best of our knowledge, this study presents the first work to examine the implications and potential of CL for enhancing gate detection performance in drone racing.

\subsubsection{Learning pipeline}
To simulate the incremental learning process, we train a model on a sequence of tasks, each representing a batch of data gathered from an environment with a unique level of illumination, illustrated in Fig.~\ref{fig:light_intensity}. Specifically, the $t$-th task represents as $\mathcal{D}^t=\{(I_i^t, M_i^t)\}_{i=1}^{N_t}$, where $N_t$ is the number of training samples. A model $f$ is designed to learn these tasks sequentially with no access to the training data from previous or future tasks. The training objective for each task is to detect the gate's position and to predict the distance and orientation between the gate and the drone. Specifically, this setup forms a Domain-Incremental Learning \cite{van2019three, van2022three} scenario, where the tasks share the same label space (identical regression targets) but differ in their input distributions (varied illumination levels). After learning the entire task sequence, the model is evaluated on the aggregated test sets from all the tasks to obtain its final overall performance.

\subsubsection{Continual learning methods}

To investigate the effect of CL, our study employs four methods grounded in two prevalent strategies. The first strategy is \textbf{Knowledge Distillation~(KD)}~\cite{hinton2015distilling}, which regularizes the student model to mimic the behaviour of a previously saved teacher. Specifically, the teacher $\hat{f}$ is initialized as a frozen copy of the model $f$ upon learning the last task. During the training of a new task, the same inputs are passed to both $f$ and $\hat{f}$ to calculate a KD loss $\mathcal{L}_{KD}$, which reflects the discrepancy between some product of the two models. By minimizing the discrepancy, knowledge acquired from previous tasks can be somewhat preserved within the current student model. Consequently, the objective function during incremental tasks is augmented with a KD loss, serving as a regularization term to preserve historical knowledge. The learning objective is formalized as follows:

\begin{equation}
    \label{eq:kd}
    \mathcal{L} = \mathcal{L}_{gate} + \lambda \mathcal{L}_{KD}.
\end{equation}
Distinct methodologies are differentiated by their respective approaches to computing $\mathcal{L}_{KD}$. We introduce two representative methods from \cite{smith2023closer}, each focusing on a different type of the model's product for distillation purposes.

\textbf{Prediction Distillation (PredKD)}: This method distils knowledge in the prediction space, specifically the outputs of the teacher and student models. It is a common practice in CL problems for classification \cite{li2017learning, rebuffi2017icarl}. Adapting this approach to our gate detection task, we redefine the KD loss as follows: 


\begin{equation}
    \label{eq:kd}
    \mathcal{L}_{KD} = \mathcal{L}_{gate}(f(I), \hat{f}(I)).
\end{equation}

\noindent This term computes the gate detection error between the output tensors of the student and teacher. During practical implementation, we clamp any negative values to zero within the confidence values in the teacher's output $\hat{f}(I)$. These modified confidence values are subsequently used to substitute the operators ${\mathds{1}}_{ij}^{\text{obj}}$ in the calculation.

\textbf{Feature Distillation (FeatKD)}: Contrary to PredKD, FeatKD focuses on the intermediate features generated by different layers of the model. A representative approach is to regularize the feature shifts with L2-distance\cite{douillard2020podnet}. The KD loss is as follows:
\begin{equation}
    \label{eq:kd}
    \mathcal{L}_{KD} = \sum_{l=1}^{L}||f^l(I) - \hat{f}^l(I)||^2_2,
\end{equation}
where $f^l$ and $\hat{f}^l$ represent the output of the $l$-th layer of the student and teacher, respectively. $L$ represent the number of layers in the feature extractor of the model.

Another strategy employed in our research is \textit{rehearsal}, which aims to reconstruct previous data distribution by saving a portion of past training data for future replay. This strategy involves storing a subset of training samples into a small memory buffer $\mathcal{M}$, in which the saved data can be revisited in subsequent tasks. The training regime is crafted to interleave new data with these retrieved historical samples, thereby updating the model in a manner that encompasses both new and past learning experiences. 

\textbf{Experience Replay (ER)}: ER is a simple but effective baseline for rehearsal  \cite{rolnick2019experience, chaudhry2019continual}. It adopts reservoir sampling~\cite{vitter1985random} to store samples and their corresponding labels in the memory buffer. During training, the saved sample label pairs are randomly retrieved for replay. Concretely, a batch of stored samples $B_\mathcal{M}$ are retrieved from the memory buffer, contributing a step of parameter update together with the batch of new data $B$.

\textbf{Dark Experience Replay (DER)}: Similar to ER, DER \cite{buzzega2020dark} follows the same two-batch regime and also uses reservoir sampling and random retrieval to manage the memory buffer. Differently, instead of saving the ground truth labels, DER stores the outputs on the previously optimized model as targets of saved samples. Such outputs, known as "dark knowledge",  encapsulate the knowledge from the optimized teacher model. Therefore, DER combines rehearsal with knowledge distillation in a way that distils in the prediction space with saved memory samples.  The original paper proposes another variant of DER termed \textbf{DER++}, which combines the ideas of ER and DER. Specifically, DER++ saves both ground truth and dark knowledge for a saved sample and uses both information types as labels during replay. 


\section{Experimental Results}
\label{sec:results}

The miniature quadrotor drone is equipped with a monocular, high-frequency Flir Blackfly RGB camera and an onboard NVIDIA Jetson TX2 computer. The fish-eye lens of the camera provides a wide field of view ($140\si{\degree}$) to ensure the observability of the next gate in the race. To create racing tracks, a set of four identical square gates -- each with internal dimensions of $1.5 \times 1.5\si{m}$ and different heights -- are placed in different layouts inside a safety-caged flying arena at Aarhus University. To provide the ground truth of the poses of the gates and the drone, we use a state-of-the-art Vicon motion capture system with 16 cameras and sub-millimeter accuracy.

 We use PencilNet~\cite{Huy2022RAL} as the backbone of the model. When learning a new task, the model is trained for 100 epochs using an Adam optimizer and a batch size of 32. The learning rate is initially set at 0.1 and subsequently reduced by a factor of 0.1 after the 5th and 10th epochs. To mitigate overfitting, an early stopping with patience=5 is used, which is triggered based on the validation loss calculated on a split validation set from the current task's training data. The memory budget for ER and DER is set to 200. Parameters remain the same as the above for the offline training. Each experiment is repeated 5 times using different random seeds.

The performances of the perception models are bench-marked using the following metrics. To measure accuracy, absolute errors~(MAE) are calculated for each testing sample by comparing the ground-truth information and the predictions of the perception model on the gate's center on the image plane~($E_c$), distance to the gate~($E_d$), and orientation of the gate~($E_{\theta}$) relative to the drone's body frame as follows:
\begin{equation}
\centering
\begin{cases}
         E_{c}  =  {
         \frac{1}{N} \sum_{i = 1}^{R}
            \sum_{j = 1}^{C}  
    \left( |\hat{x}_{ij} - x_{ij}| + |\hat{y}_{ij} - y_{ij}| \right)} \\
     E_{d}  = {
            \frac{1}{N} \sum_{i = 1}^{R}
                \sum_{j = 1}^{C}  
       |\hat{d}_{ij} - d_{ij}|} \\
 E_{\theta} =  {
            \frac{1}{N} \sum_{i = 1}^{R}
                \sum_{j = 1}^{C}  
       |\hat{\theta}_{ij} - \theta_{ij}|}     
    \end{cases},
    \label{eq:metrics}
\end{equation}
where $N$ is the number of test samples. 
For precision, a metric called average precision (AP) is utilized:

\begin{equation}
\centering
\begin{cases}
    P(n) = \frac{TP(n)}{TP(n) + FP(n)} \\
    R(n) = \frac{TP(n)}{TP(n) + FN(n)} \\
    AP = \sum_n(R(n) - R(n-1)) P(n)\\

\end{cases},
    \label{eq:precision}
\end{equation}
where $TP(n)$, $FP(n)$ and $FN(n)$ are the number of true positives, false positives and false negatives prediction of the gate object when using the threshold $n$ of the confidence value. The annotation $n$ is the current confident threshold, while $n-1$ denotes the previous threshold.

\subsection{Incremental learning from brighter to darker environments}














We first focus on the \textbf{Bright-to-Dark} setup, where the tasks are sequentially arranged according to \textit{decreasing} levels of illumination strength. Specifically, the illumination levels for the five tasks are set at $\{100,~50,~40,~20,~10\}$. This setup mirrors a common real-world scenario, where a model initially pretrained in a well-illuminated laboratory, is subsequently tasked with adapting to other environments with progressively diminishing lighting conditions. For comparison, we employ a naive sequential learning baseline (Naive) which finetunes the model without using any CL techniques, and the CL methods introduced in Section \ref{sec:method}. Besides the incremental learning pipeline, we also present the results of offline training~(Offline), i.e., the model is trained on the data of all the tasks jointly. The experiment results are shown in Table \ref{tbl:bright_to_dark}.

\begin{table}[!b]
\caption{Evaluation metrics on all test data for training sequence from brighter to darker.
\label{tbl:bright_to_dark}}
\begin{center}
\begin{tabular}{|c|cccc|}\hline
\textbf{Method} & $E_{c}(\downarrow)$ & $E_{d}(\downarrow)$ & $E_{\theta}(\downarrow)$ & AP($\uparrow$) \\

\hline
\hline

\textcolor{gray}{Offline} 

& \textcolor{gray}{0.022\tiny{$\pm$0.001}} & \textcolor{gray}{0.050\tiny{$\pm$0.003}} & \textcolor{gray}{0.025\tiny{$\pm$0.001}} & \textcolor{gray}{0.79\tiny{$\pm$0.01}}\\

\hline \hline
Naive

& 0.055\tiny{$\pm$0.009} & 0.184\tiny{$\pm$0.018} & 0.081\tiny{$\pm$0.018} & 0.50\tiny{$\pm$0.08}\\

PredKD
& 0.059\tiny{$\pm$0.012} & 0.180\tiny{$\pm$0.007} & 0.078\tiny{$\pm$0.005} & 0.47\tiny{$\pm$0.06}\\

FeatKD
& 0.056\tiny{$\pm$0.007} & 0.176\tiny{$\pm$0.011} & 0.076\tiny{$\pm$0.006} & 0.47\tiny{$\pm$0.11}\\

ER
& 0.041\tiny{$\pm$0.004} & 0.125\tiny{$\pm$0.013} & 0.055\tiny{$\pm$0.008} & 0.62\tiny{$\pm$0.07}\\

DER
& 0.042\tiny{$\pm$0.007} & 0.121\tiny{$\pm$0.020} & 0.052\tiny{$\pm$0.004} & 0.64\tiny{$\pm$0.06}\\

DER++
& \textcolor{blue}{0.040\tiny{$\pm$0.002}} & \textcolor{blue}{0.119\tiny{$\pm$0.009}} & \textcolor{blue}{0.050\tiny{$\pm$0.003}} & \textcolor{blue}{0.65\tiny{$\pm$0.03}}\\
\hline
\end{tabular}
\end{center}
\end{table}

As the upper bound of performance, Offline achieves the lowest values of errors and a high AP approaching 0.8, demonstrating the effectiveness of the training scheme for the gate detection problem with full dataset availability. However, in the incremental learning process, despite maintaining the same learning objective across tasks, there is a noticeable performance degradation compared to offline learning. Notably, Naive exhibits the most substantial performance degradation: $E_{c}$ and $E_{\theta}$ enlarge around 3 times and  $E_{d}$ enlarges even almost 4 times, and AP witnesses a drop of about 0.3. Such degradation indicates the impact of catastrophic forgetting in the domain-IL gate detection problem.

Compared with naive sequential learning, the application of CL strategies improves the final performance in general. Knowledge distillation produces modest improvements in some aspects of the results. Specifically, both KD-based methods reduce $E_{d}$ and $E_{\theta}$ slightly, though at the expense of \( E_{d} \) precision and AP. 

In contrast, rehearsal-based methods exhibit a marked enhancement across all metrics. Comparing ER and DER, we find that replaying with dark knowledge serves as a better selection than replaying with the original ground truth. Moreover, integrating both ground truth and dark knowledge for rehearsal further enhances the overall performance. 

In summary, the variation of illumination strength across tasks causes catastrophic forgetting, and using the CL strategy, especially rehearsal-based methods can mitigate this issue.

\subsection{Incremental learning from darker to brighter environments}

In this section, we investigate a more challenging \textbf{Dark-to-Bright} scenario that incrementally trains the model from darker to brighter environments. In contrast to the last section, the illumination levels for the five tasks are set at $\{10,~20,~40,~50,~100\}$. The exploration of this scenario shows the influence of task order or curriculum on CL performance. The experiment results are shown in Table \ref{tbl:dark_to_bright}.

















\begin{table}[!b]
\caption{Evaluation metrics on all test data for training sequence from darker to brighter.}
\label{tbl:dark_to_bright}
\begin{center}
\begin{tabular}{|c|cccc|}\hline
\textbf{Method} & $E_{c}(\downarrow)$ & $E_{d}(\downarrow)$ & $E_{\theta}(\downarrow)$ & AP($\uparrow$) \\

\hline
\hline

\textcolor{gray}{Offline}

& \textcolor{gray}{0.022\tiny{$\pm$0.001}} & \textcolor{gray}{0.050\tiny{$\pm$0.003}} & \textcolor{gray}{0.025\tiny{$\pm$0.001}} & \textcolor{gray}{0.79\tiny{$\pm$0.01}}\\

\hline \hline
Naive
& 0.061\tiny{$\pm$0.005} & 0.292\tiny{$\pm$0.068} & 0.057\tiny{$\pm$0.005} & 0.47\tiny{$\pm$0.03}\\

PredKD
& 0.052\tiny{$\pm$0.003} & \textcolor{blue}{0.159\tiny{$\pm$0.003}} & 0.063\tiny{$\pm$0.004} & 0.50\tiny{$\pm$0.04}\\

FeatKD
& 0.060\tiny{$\pm$0.008} & 0.287\tiny{$\pm$0.061} & 0.063\tiny{$\pm$0.017} & 0.44\tiny{$\pm$0.02}\\

ER

& 0.056\tiny{$\pm$0.009} & 0.244\tiny{$\pm$0.048} & 0.060\tiny{$\pm$0.007} & 0.47\tiny{$\pm$0.08}\\

DER

& \textcolor{blue}{0.047\tiny{$\pm$0.005}} & 0.205\tiny{$\pm$0.054} & 0.053\tiny{$\pm$0.007} & 0.53\tiny{$\pm$0.05}\\

DER++

& 0.050\tiny{$\pm$0.008} & 0.185\tiny{$\pm$0.032} & \textcolor{blue}{0.050\tiny{$\pm$0.004}} & \textcolor{blue}{0.56\tiny{$\pm$0.05}}\\

\hline
\end{tabular}
\end{center}
\end{table}

Intuitively, the detection of gates in dark environments is more challenging than that in brighter environments. Thus, the model is trained on tasks with increasing levels of difficulty. Under this curriculum setup, the experiment results are generally worse than those from the Bright-to-Dark task sequence. This is evidenced by larger error margins and a reduced AP. Such find aligns with the idea of curriculum learning \cite{bengio2009curriculum}, that learning starting from easier to more complex tasks leads to better overall performance. This phenomenon can also be interpreted from the perspective of pretraining: The model trained on previous tasks functions as the pretrained model for the current task. If the previous tasks are excessively challenging, resulting in a poorly pretrained model, the transferability of this model is diminished. This, in turn, adversely affects the learning outcomes of the current task. Conversely, the model trained on easier tasks can serve as a robust pretrained model for future tasks.

Due to the difference in the curriculum setup, the performance of various methods exhibits considerable variation across the two scenarios. In this dark-to-bright sequence order, PredKD demonstrates remarkable efficiency in overcoming catastrophic forgetting, especially for $E_d$. However, FeatKD shows a limited effect and produces outcomes similar to those of Naive. Among the rehearsal-based methods, the observed trend remains consistent: DER outperforms ER, and DER++ shows further improvement over DER. In summary, the tasks ordered from easier to harder can lead to better performance, and DER++ is the most effective method regardless of the curriculum order.


\newcommand{\ThreeFiguresWidth}{0.34}
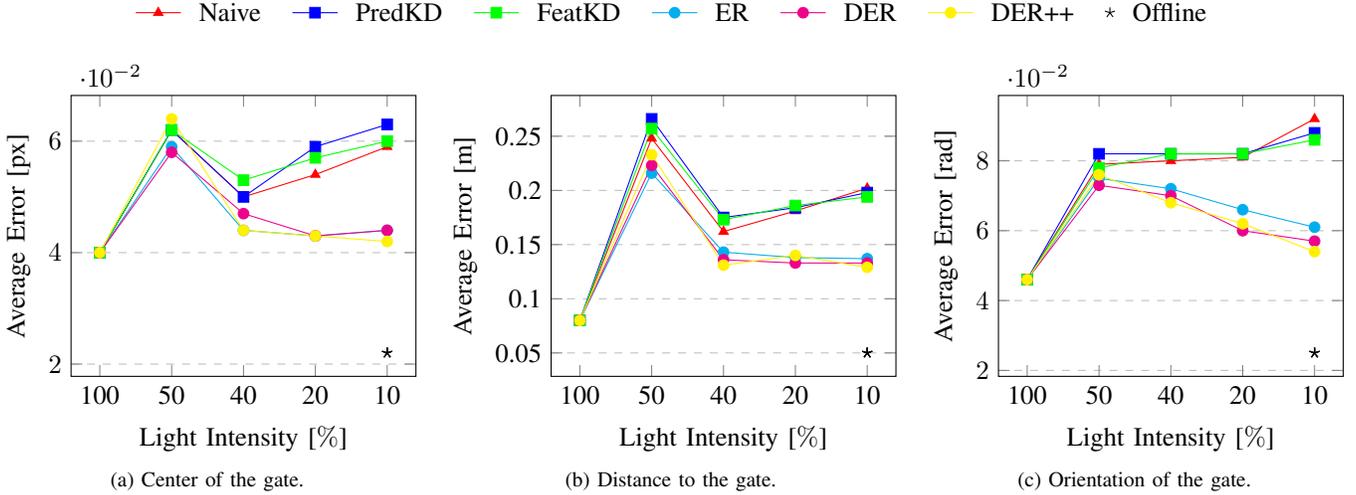
\begin{figure*}[!t]
    \pgfplotstableread[col sep=comma]{%
    light, 1, 2, 3, 4, 5
    last, 5, 5, 5, 5, 5
    'Naive', 0.04, 0.062, 0.05, 0.054, 0.059
    'Offline', 0.022, 0.022, 0.022, 0.022, 0.022
    'PredKD', 0.04, 0.062, 0.05, 0.059, 0.063
    'FeatKD', 0.04, 0.062, 0.053, 0.057, 0.06
    'ER', 0.04, 0.059, 0.044, 0.043, 0.044
    'DER', 0.04, 0.058, 0.047, 0.043, 0.044
    'DER++', 0.04,  0.064, 0.044, 0.043, 0.042
}\dataErrorCenter
\pgfplotstabletranspose[colnames from=light]\errorCenter{\dataErrorCenter}

\pgfplotstableread[col sep=comma]{%
    light, 1, 2, 3, 4, 5
    last, 5, 5, 5, 5, 5
    'Naive', 0.08, 0.248, 0.162, 0.181, 0.202
    'Offline',  0.05, 0.05, 0.05, 0.05, 0.05
    'PredKD', 0.08, 0.266, 0.175, 0.184, 0.198
    'FeatKD', 0.08, 0.257, 0.173, 0.186, 0.194
    'ER', 0.08, 0.216, 0.143, 0.138, 0.137
    'DER', 0.08,  0.223, 0.136, 0.133, 0.133
    'DER++', 0.08,  0.233, 0.131, 0.14,  0.129
}\dataErrorDistance
\pgfplotstabletranspose[colnames from=light]\errorDistance{\dataErrorDistance}

\pgfplotstableread[col sep=comma]{%
    light, 1, 2, 3, 4, 5
    last, 5, 5, 5, 5, 5
    'Naive', 0.046, 0.079, 0.08, 0.081, 0.092
    'Offline', 0.025,0.025,0.025,0.025,0.025
    'PredKD', 0.046, 0.082, 0.082, 0.082, 0.088
    'FeatKD', 0.046, 0.078, 0.082, 0.082, 0.086
    'ER', 0.046, 0.075, 0.072, 0.066, 0.061
    'DER', 0.046, 0.073, 0.07,  0.06,  0.057
    'DER++', 0.046, 0.076, 0.068, 0.062, 0.054
}\dataErrorOrientation
\pgfplotstabletranspose[colnames from=light]\errorOrientation{\dataErrorOrientation}

    \centering
    \begin{tikzpicture}
        \matrix[matrix of nodes]
        {
            \ref{legend:naive} & Naive & [5pt]
            \ref{legend:predkd} & PredKD & [5pt]
            \ref{legend:featkd} & FeatKD & [5pt]
            \ref{legend:er} & ER & [5pt]
            \ref{legend:der} & DER & [5pt]
            \ref{legend:derpp} & DER++ & [5pt]
            \ref{legend:offline} & Offline \\
        };
        
    \end{tikzpicture}
    
    \subfloat[Center of the gate.]{
        \begin{tikzpicture}
            \begin{axis}[
                xlabel={Light Intensity [$\%$]},
                ylabel={Average Error [px]},
                ylabel near ticks,
                xtick={1,2,3,4,5},
                xticklabels={100,50,40,20,10},
                width=\ThreeFiguresWidth\textwidth,
                ymajorgrids=true,
                grid style=dashed
            ]

            \addplot[color=red, mark=triangle*] table[y='Naive'] {\errorCenter}; \label{legend:naive}
            \addplot[color=blue, mark=square*] table[y='PredKD'] {\errorCenter}; \label{legend:predkd}
            \addplot[color=green, mark=square*] table[y='FeatKD'] {\errorCenter}; \label{legend:featkd}
            \addplot[color=cyan, mark=*] table[y='ER'] {\errorCenter}; \label{legend:er}
            \addplot[color=magenta, mark=*] table[y='DER'] {\errorCenter}; \label{legend:der}
            \addplot[color=yellow, mark=*] table[y='DER++'] {\errorCenter}; \label{legend:derpp}
            \addplot[color=black, mark=star, only marks] table[x=last,y='Offline'] {\errorCenter}; \label{legend:offline}
            \end{axis}
        \end{tikzpicture}
        \label{fig:b2d_center}}
    \hfill
    \subfloat[Distance to the gate.]{
        \begin{tikzpicture}
            \begin{axis}[
                xlabel={Light Intensity [$\%$]},
                ylabel={Average Error [m]},
                ylabel near ticks,
                xtick={1,2,3,4,5},
                xticklabels={100,50,40,20,10},
                ytick={0.05,0.1,0.15,0.2,0.25},
                yticklabels={0.05,0.1,0.15,0.2,0.25},
                width=\ThreeFiguresWidth\textwidth,
                ymajorgrids=true,
                grid style=dashed
            ]
            
                \addplot[color=red, mark=triangle*] table[y='Naive'] {\errorDistance};
                \addplot[color=blue, mark=square*] table[y='PredKD'] {\errorDistance};
                \addplot[color=green, mark=square*] table[y='FeatKD'] {\errorDistance};
                \addplot[color=cyan, mark=*] table[y='ER'] {\errorDistance};
                \addplot[color=magenta, mark=*] table[y='DER'] {\errorDistance};
                \addplot[color=yellow, mark=*] table[y='DER++'] {\errorDistance};
                \addplot[color=black, mark=star] table[x=last,y='Offline'] {\errorDistance};
            \end{axis}
        \end{tikzpicture}
        \label{fig:b2d_distance}}
    \hfill
    \subfloat[Orientation of the gate.]{
        \begin{tikzpicture}
            \begin{axis}[
                xlabel={Light Intensity [$\%$]},
                ylabel={Average Error [rad]},
                ylabel near ticks,
                xtick={1,2,3,4,5},
                xticklabels={100,50,40,20,10},
                width=\ThreeFiguresWidth\textwidth,
                ymajorgrids=true,
                grid style=dashed
            ]
            
                \addplot[color=red, mark=triangle*] table[y='Naive'] {\errorOrientation};
                \addplot[color=blue, mark=square*] table[y='PredKD'] {\errorOrientation};
                \addplot[color=green, mark=square*] table[y='FeatKD'] {\errorOrientation};
                \addplot[color=cyan, mark=*] table[y='ER'] {\errorOrientation};
                \addplot[color=magenta, mark=*] table[y='DER'] {\errorOrientation};
                \addplot[color=yellow, mark=*] table[y='DER++'] {\errorOrientation};
                \addplot[color=black, mark=star] table[x=last,y='Offline'] {\errorOrientation};
            \end{axis}
        \end{tikzpicture}
        \label{fig:b2d_orientation}}
    \caption{Evolution of Average Error for \textbf{bright-to-dark} training sequence.}
    \label{fig:error_b2d}
\end{figure*}

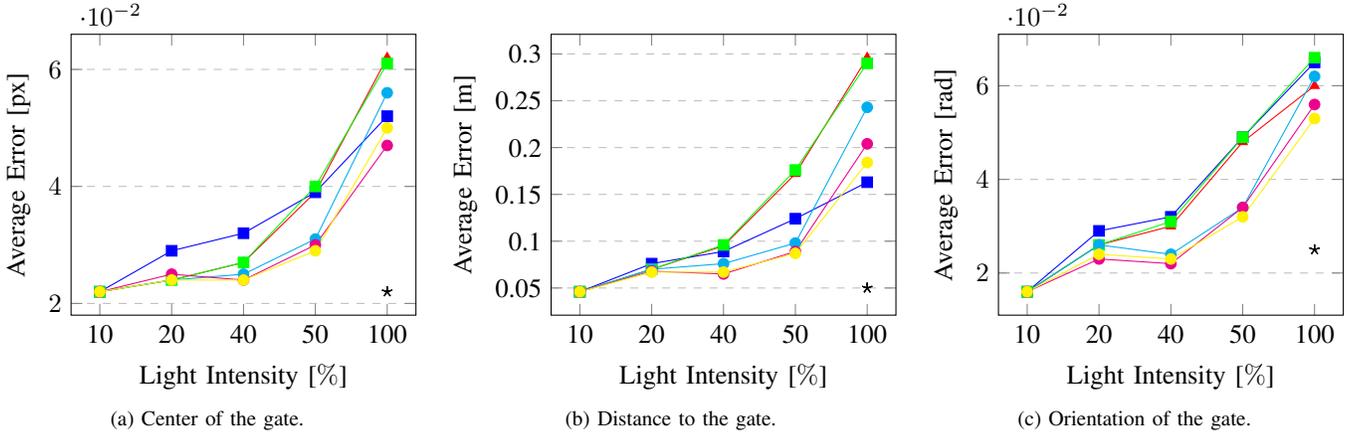
\begin{figure*}[!t]
    \pgfplotstableread[col sep=comma]{%
    light, 1, 2, 3, 4, 5
    last, 5, 5, 5, 5, 5
    'Naive', 0.022, 0.024, 0.027, 0.039, 0.062
    'Offline', 0.022,0.022,0.022,0.022,0.022
    'PredKD', 0.022, 0.029, 0.032, 0.039, 0.052
    'FeatKD', 0.022, 0.024, 0.027, 0.04,  0.061
    'ER', 0.022, 0.024, 0.025, 0.031, 0.056
    'DER', 0.022, 0.025, 0.024, 0.03,  0.047
    'DER++', 0.022, 0.024, 0.024, 0.029, 0.05
}\dataErrorCenter
\pgfplotstabletranspose[colnames from=light]\errorCenter{\dataErrorCenter}

\pgfplotstableread[col sep=comma]{%
    light, 1, 2, 3, 4, 5
    last, 5, 5, 5, 5, 5
    'Naive', 0.046, 0.07,  0.095, 0.172, 0.296
    'Offline',  0.05,0.05,0.05,0.05,0.05
    'PredKD', 0.046, 0.076, 0.089, 0.124, 0.163
    'FeatKD', 0.046, 0.07,  0.096, 0.176, 0.29
    'ER', 0.046, 0.07,  0.076, 0.098, 0.243
    'DER', 0.046, 0.068, 0.065, 0.089, 0.204
    'DER++', 0.046, 0.067, 0.067, 0.087, 0.184
}\dataErrorDistance
\pgfplotstabletranspose[colnames from=light]\errorDistance{\dataErrorDistance}

\pgfplotstableread[col sep=comma]{%
    light, 1, 2, 3, 4, 5
    last, 5, 5, 5, 5, 5
    'Naive', 0.016, 0.026, 0.03,  0.048, 0.06
    'Offline', 0.025,0.025,0.025,0.025,0.025
    'PredKD', 0.016, 0.029, 0.032, 0.049, 0.065
    'FeatKD', 0.016, 0.026, 0.031, 0.049, 0.066
    'ER', 0.016, 0.026, 0.024, 0.034, 0.062
    'DER', 0.016, 0.023, 0.022, 0.034, 0.056
    'DER++', 0.016, 0.024, 0.023, 0.032, 0.053
}\dataErrorOrientation
\pgfplotstabletranspose[colnames from=light]\errorOrientation{\dataErrorOrientation}

        
    
    \subfloat[Center of the gate.]{
        \begin{tikzpicture}
            \begin{axis}[
                xlabel={Light Intensity [$\%$]},
                ylabel={Average Error [px]},
                ylabel near ticks,
                xtick={1,2,3,4,5},
                xticklabels={10,20,40,50,100},
                width=\ThreeFiguresWidth\textwidth,
                ymajorgrids=true,
                grid style=dashed
            ]

            \addplot[color=red, mark=triangle*] table[y='Naive'] {\errorCenter}; \label{legend:naive}
            \addplot[color=blue, mark=square*] table[y='PredKD'] {\errorCenter}; \label{legend:predkd}
            \addplot[color=green, mark=square*] table[y='FeatKD'] {\errorCenter}; \label{legend:featkd}
            \addplot[color=cyan, mark=*] table[y='ER'] {\errorCenter}; \label{legend:er}
            \addplot[color=magenta, mark=*] table[y='DER'] {\errorCenter}; \label{legend:der}
            \addplot[color=yellow, mark=*] table[y='DER++'] {\errorCenter}; \label{legend:derpp}
            \addplot[color=black, mark=star, only marks] table[x=last,y='Offline'] {\errorCenter}; \label{legend:offline}
            \end{axis}
        \end{tikzpicture}
        \label{fig:b2d_center}}
    \hfill
    \subfloat[Distance to the gate.]{
        \begin{tikzpicture}
            \begin{axis}[
                xlabel={Light Intensity [$\%$]},
                ylabel={Average Error [m]},
                ylabel near ticks,
                xtick={1,2,3,4,5},
                xticklabels={10,20,40,50,100},
                ytick={0.05,0.1,0.15,0.2, 0.25, 0.3, 0.35},
                yticklabels={0.05,0.1,0.15,0.2, 0.25, 0.3, 0.35},
                width=\ThreeFiguresWidth\textwidth,
                ymajorgrids=true,
                grid style=dashed
            ]
            
                \addplot[color=red, mark=triangle*] table[y='Naive'] {\errorDistance};
                \addplot[color=blue, mark=square*] table[y='PredKD'] {\errorDistance};
                \addplot[color=green, mark=square*] table[y='FeatKD'] {\errorDistance};
                \addplot[color=cyan, mark=*] table[y='ER'] {\errorDistance};
                \addplot[color=magenta, mark=*] table[y='DER'] {\errorDistance};
                \addplot[color=yellow, mark=*] table[y='DER++'] {\errorDistance};
                \addplot[color=black, mark=star] table[x=last,y='Offline'] {\errorDistance};
            \end{axis}
        \end{tikzpicture}
        \label{fig:b2d_distance}}
    \hfill
    \subfloat[Orientation of the gate.]{
        \begin{tikzpicture}
            \begin{axis}[
                xlabel={Light Intensity [$\%$]},
                ylabel={Average Error [rad]},
                ylabel near ticks,
                xtick={1,2,3,4,5},
                xticklabels={10,20,40,50,100},
                width=\ThreeFiguresWidth\textwidth,
                ymajorgrids=true,
                grid style=dashed
            ]
            
                \addplot[color=red, mark=triangle*] table[y='Naive'] {\errorOrientation};
                \addplot[color=blue, mark=square*] table[y='PredKD'] {\errorOrientation};
                \addplot[color=green, mark=square*] table[y='FeatKD'] {\errorOrientation};
                \addplot[color=cyan, mark=*] table[y='ER'] {\errorOrientation};
                \addplot[color=magenta, mark=*] table[y='DER'] {\errorOrientation};
                \addplot[color=yellow, mark=*] table[y='DER++'] {\errorOrientation};
                \addplot[color=black, mark=star] table[x=last,y='Offline'] {\errorOrientation};
            \end{axis}
        \end{tikzpicture}
        \label{fig:b2d_orientation}}
    \caption{Evolution of Average Error for \textbf{dark-to-bright} training sequence.}
    \label{fig:error_d2b}
\end{figure*}

\subsection{Analysis of knowledge transfer in learning process}
\label{subsec: CL analysis}
In this section, we further analyze the knowledge transfer in the incremental learning process in the two scenarios. The enhancement of overall performance in CL methods can be primarily attributed to two pivotal aspects: the efficiency in assimilating new tasks, demonstrating the model's \textit{plasticity}, and the retention of previously acquired knowledge, showcasing the model's \textit{stability}. To investigate the sources of the performance gains, we quantify these two characteristics by adapting and applying two specialized CL metrics for our regression context.  

Let $T$ denote the total number of tasks in the sequence, and let $e_{i,j}$ represent the testing error on task $j$ after learning task $i$, where $j \leq i$. Learning error~(LE) reflects the overall effect of a CL method on the model's plasticity. It is the average error for each task directly after it is learned, formally expressed as $\text{LE} = \frac{1}{T} \sum_{i=1}^{T} e_{i,i}$. Conversely, Forgetting Measure~(FM) serves as an indicator of stability, which quantifies how much previously learned knowledge is diminished due to the incorporation of new tasks. FM after learning the entire task sequence is defined as $\text{FM} = \frac{1}{T-1}\sum_{j=1}^{T-1} f_{T,j}$, where $f_{T,j}= e_{T, j} - \min_{k \in \{1,...,T-1\}} (e_{k,j}), j<i$  represents the increase in error for task $j$ over its minimum past error after the learning of the whole task sequence. 

\begin{table}[!b]
    \centering
    \caption{The results of Learning Error and Forgetting Metric on the two incremental task sequences.}
    \resizebox{0.98\columnwidth}{!}{
    \begin{NiceTabular}{|c|c|ccc|ccc|}
        \toprule
        \multirow{2}{*}{\textbf{Metric}} & \multirow{2}{*}{\textbf{Method}} & \multicolumn{3}{c}{\textbf{Bright-to-Dark}} & \multicolumn{3}{c}{\textbf{Dark-to-Bright}} \\ \cmidrule{3-8}
         & & $E_{c}$ & $E_{d}$ & $E_{\theta}$ & $E_{c}$ & $E_{d}$ & $E_{\theta}$ \\
        
        \midrule
        \multirow{6}{*}{\textbf{LE $(\downarrow)$}} & Naive & 0.023 & \textcolor{blue}{0.047} & \textcolor{blue}{0.023} & \textcolor{blue}{0.021} & \textcolor{blue}{0.048} & \textcolor{blue}{0.022} \\
                        ~ & PredKD & 0.023 & 0.049 & \textcolor{blue}{0.023} & 0.042 & 0.137 & 0.056 \\
                        ~ & FeatKD & \textcolor{blue}{0.022} & \textcolor{blue}{0.047} & \textcolor{blue}{0.023} & 0.022 & 0.050 & 0.023 \\
                        ~ & ER & 0.024 & 0.050 & 0.026 & 0.023 & 0.051 & 0.025 \\
                        ~ & DER & 0.025 & 0.052 & 0.026 & 0.024 & 0.055 & 0.025 \\
                        ~ & DER++ & 0.026 & 0.056 & 0.027 & 0.023 & 0.050 & 0.024 \\
        \midrule

        \multirow{6}{*}{\textbf{FM $(\downarrow)$}} & Naive & 0.061 & 0.227 & 0.101 & 0.065 & 0.343 & 0.063 \\
                ~ & PredKD & 0.065 & 0.220 & 0.096 & \textcolor{blue}{0.042} & \textcolor{blue}{0.132} & \textcolor{blue}{0.052} \\
                ~ & FeatKD & 0.062 & 0.216 & 0.094 & 0.063 & 0.334 & 0.068 \\
                ~ & ER & 0.041 & 0.145 & 0.062 & 0.057 & 0.275 & 0.063 \\
                ~ & DER & 0.041 & 0.139 & 0.056 & 0.046 & 0.224 & 0.055 \\
                ~ & DER++ & \textcolor{blue}{0.038} & \textcolor{blue}{0.133} & \textcolor{blue}{0.053} & 0.050 & 0.203 & 0.053 \\

        \bottomrule
    \end{NiceTabular}}
    \label{tbl:cl_metrics}
\end{table}

Table \ref{tbl:cl_metrics} presents LE and FM metrics computed in two experimental scenarios. In both scenarios, CL methods predominantly yield LE values that are comparable to or marginally higher than the baseline while simultaneously maintaining consistently lower FM values. This pattern suggests that the enhancement observed in the final outcomes is primarily due to the preservation of previous knowledge. Moreover, the influence of old knowledge retention appears to slightly impede the learning of new tasks. Nevertheless, the improvement of stability is more pronounced than the nominal decrement in plasticity, resulting in an overall improvement in performance.

These results also offer insight into the efficacy of different CL methods. Notably, DER++ and DER outperform ER in terms of stability retention, as reflected by their lower FM. It is also noteworthy that PredKD exhibits robust competitiveness in the Dark-to-Bright scenario, as it achieves a lower LE for $E_d$. This suggests that the KD regularization employed by PredKD may particularly enhance the learning efficiency for distance prediction in new tasks.

In addition to the aforementioned metrics, we also present the evolution of average error throughout the incremental learning process. The average error indicates the model’s proficiency in assimilating and retaining knowledge across all tasks encountered up to a given task $i$, which is expressed as $\text{AE}_i = \frac{1}{i} \sum_{j=1}^{i} e_{i,j}$. The evolution of average error on Bright-to-Dark and Dark-to-Bright scenarios are depicted in Figs.~\ref{fig:error_b2d}~and~\ref{fig:error_d2b}, respectively.

The evolution of average errors exhibits distinct patterns between the two task sequences. In the Dark-to-Bright scenario, there is a simpler general trend wherein the average errors escalate as additional tasks are introduced. This increment can be rationalized by the mounting forgetfulness inherent to the learning of new tasks, which gradually degrades the model's performance over time. Contrarily, the Bright-to-Dark sequence does not manifest a monotonous trend. Rather, the average errors drastically surge upon learning the second task, followed by a reduction after learning the third task. We posit that this fluctuation stems from the varying degrees of similarity between the consecutive tasks. A pronounced change in illumination, particularly from 100\% to 50\%, between the first two tasks may induce significant distribution shifts, and result in severe forgetting. Subsequently, the proximity in illumination levels between the second and third tasks appears to reduce task dissimilarity, facilitating learning and yielding an improved average error. In conclusion, the evolution of average error underscores the pivotal role of task similarity and emphasizes the need for careful consideration of task similarity when learning from a sequence of tasks.


\section{Conclusions and Future Work}
\label{sec:conclusions}


In this study, we explore the effectiveness of various CL techniques in enhancing gate detection for autonomous drone racing, with a specific focus on coping with dynamic lighting conditions. Our findings reveal that fluctuating illumination strengths across tasks can lead to catastrophic forgetting, significantly impeding the neural network's ability to retain previously learned information. However, the deployment of CL strategies, particularly those based on rehearsal methods, has proven to be a viable solution to mitigate these adverse effects. Moreover, the order of tasks plays a crucial role in the overall performance of the CL system. Organizing tasks from simpler to more complex not only facilitates yields superior outcomes. Among the various methods evaluated, DER++ emerges as the standout technique, delivering robust performance irrespective of the task ordering. Despite the notable improvements in performance attributed to the adoption of CL strategies, all methods are subject to the inherent stability-plasticity dilemma. This phenomenon underscores the ongoing challenge of balancing the need for the model to remain adaptable to new information while retaining previously acquired knowledge. Our study not only demonstrates the potential of continual learning to effectively navigate the complexities of real-world perception tasks in robotics but also highlights critical considerations for optimizing CL systems for better adaptability and knowledge retention under varying environmental conditions.

In the future, the proposed framework can be extended to online continual learning to be able to adapt in real-time to new environmental conditions.


\bibliographystyle{IEEEtran}
\bibliography{bibliography}

\end{document}